# Synthesizing Scientific Summaries: An Extractive and Abstractive approach

Grishma Sharma*, Aditi Paretkar* and Deepak Sharma

*Abstract*—The availability of a vast array of research papers in any area of study, necessitates the need of automated summarisation systems that can present the key research conducted and their corresponding findings. Scientific paper summarisation is a challenging task for various reasons including token length limits in modern transformer models and corresponding memory and compute requirements for long text. A significant amount of work has been conducted in this area, with approaches that modify the attention mechanisms of existing transformer models and others that utilise discourse information to capture long range dependencies in research papers. In this paper, we propose a hybrid methodology for research paper summarisation which incorporates an extractive and abstractive approach. We use the extractive approach to capture the key findings of research, and pair it with the introduction of the paper which captures the motivation for research. We use two models based on unsupervised learning for the extraction stage and two transformer language models, resulting in four combinations for our hybrid approach. The performances of the models are evaluated on three metrics and we present our findings in this paper. We find that using certain combinations of hyper parameters, it is possible for automated summarisation systems to exceed the abstractiveness of summaries written by humans. Finally, we state our future scope of research in extending this methodology to summarisation of generalised long documents.

**Impact Statement:** Students or researchers working in any domain of research are required to read through many related works in their area of study, before they can start working on something novel. Usually, the initial phase of going through many related works is time consuming and tedious. The time and energy spent in going through related works can instead be invested in designing their new approach, if only shorter and concise summaries of the work are made available to them. Our work aims to address this, by devising a new method for summarization, that can potentially help such people. By specifically fine-tuning to a dataset of scientific papers, we aim to improve the performances of our models on scientific document summarization.

## I. INTRODUCTION

Text summarization is the process of creating an abridged version of a text. The abridged version, generally referred to as the summary of the text, is shorter in length than the original text and encapsulates the principal points. With the growth in both the length of text and the information overflow related to a particular topic, the availability of shorter and more concise summaries of text makes it easier to skim through many resources available online, decide which information is relevant to a particular task, and subsequently gather the relevant data required for further processing. This need becomes especially apparent in the case of the summarization of scientific papers. Scientific work, such as that published in journal articles or conference proceedings, contains many sections of text with other information such as diagrams, tables, and flowcharts. To get a condensed idea of the work conducted and relevant results achieved, a concise summary of the paper can prove to be useful.

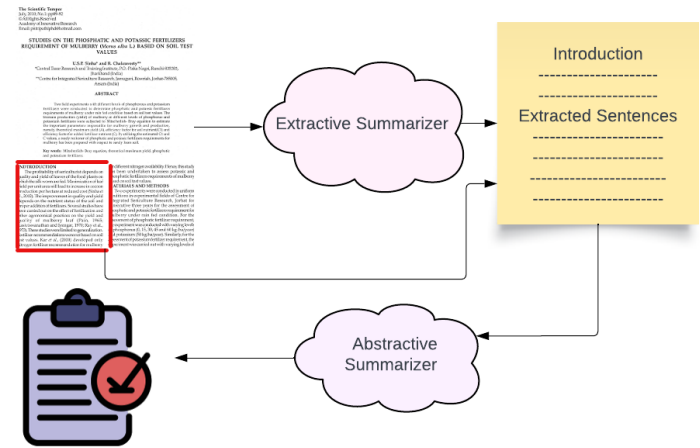

Fig.1 This diagram displays an overview of our approach. We extract the important sentences from a paper by extractive algorithms and the original introduction. The input to the abstractive models is structured such that the introduction comes first, followed by the extracted sentences, which are used to generate the final abstractive summary.

Much work has been conducted in long document summarization. With the development of transformer models, specific changes in their architecture or variations of pre-training data, along with fine-tuning for specific downstream tasks, have proven to be useful for long document summarization [1]. Summarization, which is broadly classified into extractive and abstractive techniques, has been automated largely by various algorithms and by transformer language models.

In this paper, we propose a novel approach to summarizing scientific articles by following three steps: First, we extract the key sentences from a scientific text using three different algorithms that use a reinforcement learning approach to perform extractive summarization. Next, we use the extracted summaries and concatenate them with the introduction of the paper as the input to a few transformer models, providing the actual abstract of the paper as the ground truth during a "fine-tuning" stage. We contend that the introduction to a paper conveys the motivation to perform research along with a short summary of the technique used. This, coupled with the key sentences, should provide the context sufficient to summarize the document. Finally, we perform an inference step using a fraction of the data given during training to the model and save the summaries generated. This inference step is performed for several combinations of hyperparameters, and the results are noted.

Prior works on hybrid approaches to long document summarisation include the work by [2] that also utilizes two unsupervised models for the extractive step. For the abstraction, they introduce an automated method for labeled data creation which randomly samples 3 sentences from the source document as the target and selects sentences from the source document with highest similarity to the target sentences. Another approach by [3] considers extreme summarization of scientific documents where they summarize the long scientific documents and create a single line as a summary. Their approach also requires expert background knowledge in performing a task. In contrast, our approach tries to create a holistic summary encompassing information across various sections of a scientific paper, into a summary consisting of 5-6 lines.

We perform the final evaluation of the quality of summaries generated during inference using three different techniques: ROUGE scores, the number of unique n-grams generated in the final summary and a human-evaluation step to assess the coherence and comprehensibility, and to evaluate the degree of semantic similarity between the model generated and ground truth summaries.

The rest of the paper is structured as follows: First we provide a background of related works that have been performed for abstractive summarisation of long documents in Section 2. In Section 3, we provide a detailed overview of our methodology followed by the experimental setup details in section 4. Finally, in section 5 we discuss the results obtained by assessing them through various metrics of evaluation and conclude our paper in section 6.

## II. RELATED WORK

In this section, we aim to provide a brief overview of the work conducted so far in long-text summarisation and specifically in those areas that address the issues inherent in long-text summarisation. Before generating an abstract, performing the extractive summarisation step helps to condense the information, to retain the important information. The work by [4] utilises Restricted Boltzmann Machines in the feature enhancement step along with various heuristics to select the important sentences to form the extractive summary. Furthermore, [5] propose a hierarchical and structural ranking model, which allows user-preference based sentence selection and weight assignment to an unsupervised model for extractive summarisation of long scientific documents. For a comprehensive review of text summarisation techniques based on other strategies of categorization, we refer the reader to [6], here we aim to provide a brief review of some of the work done so far.

### A. Hybrid Approaches

Hybrid approaches to long document summarization often combine extractive and abstractive methods. One approach by [7] uses ROUGE scores to select semantically similar sentences for extraction, followed by generating abstractive summaries using a transformer language model trained from scratch. Another method proposed by [8] reframes abstractive summarization as paraphrasing, with an initial extractive step guided by reinforcement learning based on semantic overlap, followed by a paraphrase function to compress selected sentences into a summary. A third strategy by [9] employs a divide-and-conquer technique, summarizing individual sections of a document and then aggregating these summaries. In this method, each line from the target summary is matched with sentences from document sections using ROUGE-L scores, and beam search is used to generate and combine section summaries.

### B. Transformer Models

The original attention paper by [10] introduced a transformer model that replaced recurrence and convolution with multi-head self-attention and feed-forward neural networks for text processing tasks like text summarization. Since then, many transformer models have been developed. [11] explored unsupervised learning with a teacher-forcing mechanism, masking a percentage of the data during pre-training and expecting the model to predict the masked data. This approach treats training as a text-to-text task, predicting text outputs from text inputs. For summarization, particularly on the CNNDM dataset, the authors found that beam search works effectively.

### C. Addressing memory and compute requirements

Adapting pre-trained models to long scientific documents, such as research papers, presents challenges due to input token size limitations. To address this, several methodologies have been proposed. One such approach by [12] involves a top-down and bottom-up strategy to manage quadratic memory requirements: the bottom-up approach uses self-attention within a small window size to capture fine context, becoming coarser at higher layers, while the top-down approach captures long-range dependencies across larger segments. Additionally, a combination of Local Attention and Content Selection has been proposed by [13] to reduce memory requirements by using a local attention window and re-ranking sentences based on their relevance to the target summary, truncating inputs to include higher-ranked sentences within acceptable input size limits.

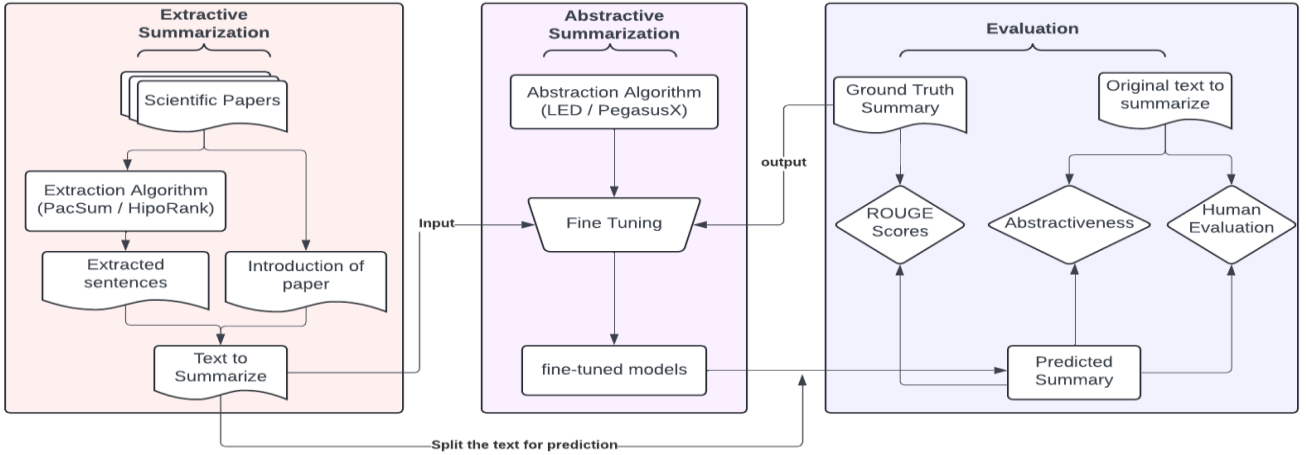

Fig. 2 Our proposed methodology

*D. Utilizing Discourse Information in Text Summarization*

In long documents, particularly scientific papers with multiple sections, understanding the main points is crucial for creating coherent summaries.[14] proposed creating Elementary Discourse Units (EDUs), which are either sentences or parts of sentences that convey specific pieces of information relevant to the discourse. These EDUs are classified as Nucleus or Satellite based on their relevance and are linked using a co-reference graph to model relationships between distantly placed but related EDUs. Another method by [15] addresses coreference resolution with a three-layer model consisting of an NLP module for preprocessing and coreference tasks, an NLU layer that utilizes word embeddings, and an NLG module for the summarization task. A third approach proposed by [16] aims to generate summaries through extractive summarization by first using a content-ranking module to rank sentences based on their importance to the summary and then building a graph representation of the document to capture the relationships and locations of these sentences within the source text.

*E. Augmenting Input Data to enhance summaries*

In [17], the authors experiment with providing "guidance signals" to models to improve the faithfulness of extractive summaries. These signals, which can be either manually created or automated, include highlighted sentences, important keywords, and relational tuples to enhance the quality of the summaries. In [18], the authors address the issue of pre-trained transformers focusing on a single key event by augmenting the BART architecture with a set of key events, using input prompting and attention masks to distinguish between source text and event texts. In [19], the authors extend the Information Bottleneck principle to reduce redundancy in the summarization of long scientific documents. They perform extractive summarization by defining correlation signals that present different views of the input document, using various state-of-the-art techniques or self-defined signals.

III. METHODOLOGY

In this section, we introduce our framework and detail the steps followed to generate the abstractive summary in our two step framework. To put it succinctly, our framework consists of two steps that consist of an extractive and abstractive step. The extractive step makes use of unsupervised algorithms for identifying the salient sentences. The reason for the choice of unsupervised algorithms is deemed necessary as supervised algorithms do not perform as expected on long document summarization pertaining to scientific articles due to lack of labeled training data and greater length of scientific articles [1]. Figure 2 illustrates the steps we have followed in our approach

*A. Extraction Step*

Our framework starts with a step that renders the xml document of a scientific article. The xml document is provided as an input to two unsupervised algorithms (PacSum and HIPORank) to generate the extractive summaries. We first provide a brief description of these two algorithms below.

**PacSum**: The **P**osition-**A**ugmented **C**entrality based **Sum**marization algorithm [20], is based on the revised notion of node centrality, where the "nodes" are taken to represent the sentences of a document. This algorithm posits that the weight given to any node should be conditioned on the position of the node, following the assumption that the relative position of a sentence (node) in a document decides its salience. Following this, the authors propose that undirected edge $e_{ij}$ in a graph between any two sentences $s_i$ and $s_j$ can be expressed as two directed edges $\lambda_1 e_{ij}$ and $\lambda_2 e_{ij}$, where $\lambda_1$ and $\lambda_2$ represent different weights for forward and backward edges. Then the centrality of a node is given by the following equation:

$$centrality(s_i) = \lambda_1 \sum_{j<i} e_{ij} + \lambda_2 \sum_{j>i} e_{ij} \quad (1)$$

**HIPORank**: The **Hi**erarchical and **Po**sitional **Rank**ing Algorithm **[**21**]** introduces the concept of sentence and section saliency by proposing that sentences closer to the boundaries of a section are more likely to be important sentences that could be selected to formulate the extractive summary of long scientific documents. This algorithm also creates bi-directional edges between sentences (which are represented by nodes) and assigns weights to them by:

(i) computing a similarity score between any two sentences

(ii) multiplying the score with a factor $\lambda_1$ or $\lambda_2$, where $\lambda_1 < \lambda_2$ depending on the distance of the sentence to the section boundary relative to the sentence that it is being compared with. The weight for intra-section edges is defined as:

$$w_{ji}^I = \begin{cases} \lambda_1 * sim(v_j^I, v_i^I), & if\ d_b(v_i^I) \geq d_b(v_j^I) \\ \lambda_2 * sim(v_j^I, v_i^I), & if\ d_b(v_i^I) < d_b(v_j^I) \end{cases} \quad (2)$$

Here, the function $d_b$ is a sentence boundary function that represents the distance of a sentence to the boundary of the section that it is contained in, and $w_{ji}^I$ represents the incoming edge to sentence $v_i$. A similar method is utilized to calculate the importance of a section relative to other sections of the document. Finally, the importance of a sentence is calculated as the weighted sum of its inter-section and intra-section centrality scores.

### B. Fine-tuning Step

We propose utilising two Transformer Language Models with a sufficiently long admissible input token length for summarizing the documents created. A short description of these is provided below:

**Pegasus-X**: This model extends the $PEGASUS_{LARGE}$ model to address the issues inherent in it for handling longer token sequences. These issues mainly stem from the fact that the pre-training of the model has been performed on shorter input tokens (limited to 512 or 2048 tokens) and fine-tuning such models on longer tokens can result in 32 times more memory consumption for encoder self-attention, in fine tuning, as compared to pretraining. The issues are addressed by introducing several architectural changes such as improvising local attention blocks to staggered local attention blocks which helps to exchange information across different blocks. The authors also experiment with various other settings such as Local and Global-Local Configurations, Positional Encoding Schemes, Scaling Encoder and Decoder Layers, Pretraining Schemes describing the results of pre-training on shorter and longer input documents and dropping cross attention for a fraction of decoder layers, for a slight performance trade-off, in favour of addressing memory requirements. [1]

**Longformer Encoder Decoder (LED)**: This model addresses the shortcomings of modern transformer architectures that incur large memory and computational costs involved in the self-attention mechanism, which scales quadratically with the size of the input sequence length. The attention pattern in this model scales linearly with the input sequence length by involving stacked layers of windows that are involved in the attention mechanism. Specifically, if the window size is $w$, each token attends to $\frac{1}{2}w$ tokens on each side of it. The windows at each layer are able to process more information than the windows at the previous layer. This has been complemented by adding a "dilated" sliding window mechanism along with incorporating global attention for a few input tokens. This global attention is symmetric, i.e. the tokens having global attention attend to all tokens across the sequence and all tokens across the sequence attend to the tokens with global attention.[11]

### C. Abstractive Summarization and Inference

After fine-tuning has been performed, we then utilise our models for abstractive summarization tasks. We perform inference on the models by using a portion of the test data (details in next section), and utilise three methodologies to evaluate the summaries generated in various aspects.

## IV. EXPERIMENTAL SETUP

### A. Dataset

We utilize the SciSummNet dataset introduced in [22]. The reason for the choice of this dataset is that it consists of around 1000 research papers, which is an appreciable number for fine-tuning tasks. According to the authors, the papers have been shortlisted as the 1000 most cited papers in the ACL Anthology Network (AAN). The authors have generated a reference summary for each paper by extracting 20 citation sentences about the paper, which give a context about the opinion of the research conducted in the paper from the authors of research articles that cite the paper. This is combined with a few sentences from the abstract to generate the reference summary of the paper.

For our proposed model, we extract the important sentences from each of the papers using both the HIPORank and PacSum models and store them. We then extract the sentences present in the introduction section from each paper and combine them with the extracted sentences, to form the input to the models during fine-tuning. The reference summaries provided by the authors are provided as the target (ground truth) to the models during fine-tuning. The data is then split into training, validation and test sets, in the ratio of 0.8, 0.1 and 0.1 respectively

### B. Details of Models

We utilise the extractive summaries from HIPORank and PacSum to train each of Pegasus-X and LED models separately, resulting in a combination of 4 models as shown in Fig. 3.

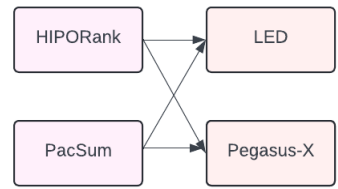

Fig. 3 Our extractive and abstractive approaches are carried by 4 combinations.

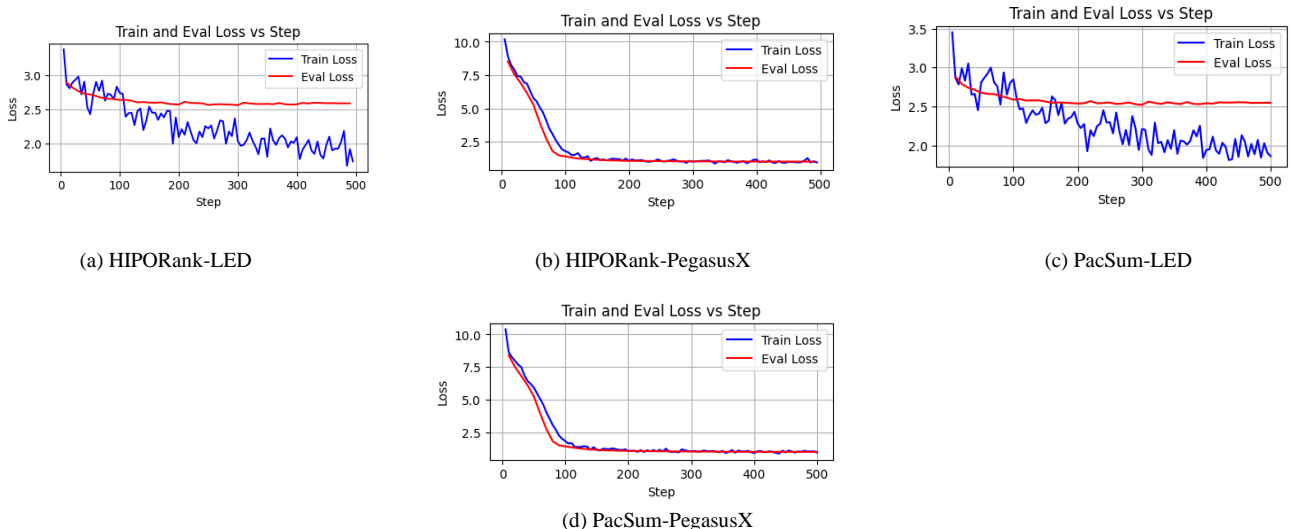

Fig. 4 Training and validation loss graphs for the models during fine-tuning

For the tokenization of data, we utilize the pre-trained AutoTokenizer from the HuggingFace library. For fine-tuning the LED model, we use AutoTokenizer from the `allenai/led-base-16384` checkpoint and for fine-tuning the Pegasus Model we utilize the Autotokenizer from the `google/pegasus-x-large` checkpoint. We set a minimum length of 4000 for the encoder input tokens and 512 for the output tokens. Henceforth, we refer to each model by [name of unsupervised model used in extraction stage – (hyphen) name of model used in abstraction stage], example HIPORank-PegasusX.

*C. Fine-Tuning Details*

To fine-tune the models, we used the L4 and A100 GPUs with a capacity of 24GB RAM and 40GB RAM respectively, and performed training over 5 epochs. A few values of the hyperparameters used during fine-tuning were `max_length = 512, min_length = 100, length_penalty = 2.0, no_repeat_ngram_size = 3, train_batch_size = 2, eval_batch_size = 2`. The graphs of training and validation losses for the combination of the four models are shown in Fig 4.

The training and validation losses for the HIPORank-LED model started at 2.861 and 2.892 for the first epoch and showed a monotonic decrease till the last epoch, approaching a value of 1.919 and 2.587 at the end of the $5^{th}$ epoch. In contrast, the HIPORank-PegasusX model started at 8.909 and 8.534 for the first epoch, decreasing rapidly and plateauing at 1.036 and 1.013, respectively at the end of the $5^{th}$ epoch. The training and validation losses for the PacSum-LED model started out at values of 2.864 and 2.881, reaching values of 2.861 and 1.546 at the end of the $5^{th}$ epoch, while the training and validation losses for the PacSum-PegasusX model started at 8.656 and 8.404, converging at values of 0.953 and 1.004 at the last epoch. The validation loss thus shows a steady trend of convergence with the training loss when PegasusX is utilized for abstractive summarization. This trend shows that although the LED model started out better than PegasusX for the abstractive summarization step, PegasusX was able to achieve a better training and validation performance over 5 epochs.

## V. RESULTS AND DISCUSSION

After fine-tuning each model, we performed inferences on each of the models by testing a combination of hyperparameters. The details of the various combinations of hyperparameters used are provided in Table I.

TABLE I
DETAILS OF HYPERPARAMETERS DURING INFERENCE

| Hyperparameter | Setting 1 | Setting 2 | Setting 3 |
|---|---|---|---|
| max_length | 512 | 512 | 512 |
| min_length | 100 | 100 | 150 |
| length_penalty | 2.0 | 1.0 | 1.0 |
| no_repeat_ngram_size | 4 | 3 | 2 |
| num_beams | 4 | 2 | 3 |
| do_sample | false | false | false |

A brief explanation of the various hyperparameters used during inference can be found below:

1. `max_length`: sets an upper limit to maximum length of generated summary
2. `min_length`: sets a lower limit to maximum length of generated summary
3. `length_penalty`: this hyperparameter controls the length of the generated sequence by penalizing the model to generating summaries that are neither overly short nor overly long.
4. `no_repeat_ngram_size`: for any value greater than 0, all n-grams of that size can occur only once
5. `num_beams`: the beam size used in beam search algorithm
6. `do_sample`: this flag determines whether to use sampling

TABLE II
ROUGE SCORES FOR DIFFERENT SETTINGS

| Model | Setting | ROUGE-1 | ROUGE-2 | ROUGE-L |
|---|---|---|---|---|
| HIPORank-LED | Setting 1 | 0.394 | 0.188 | 0.364 |
| | Setting 2 | 0.409 | **0.205** | 0.380 |
| | Setting 3 | **0.433** | 0.195 | **0.404** |
| HIPORank-PegasusX | Setting 1 | 0.374 | 0.173 | 0.343 |
| | Setting 2 | 0.371 | 0.168 | 0.340 |
| | Setting 3 | 0.369 | 0.151 | 0.339 |
| PacSum-LED | Setting 1 | 0.383 | 0.180 | 0.360 |
| | Setting 2 | 0.400 | 0.195 | 0.374 |
| | Setting 3 | 0.394 | 0.171 | 0.366 |
| PacSum-PegasusX | Setting 1 | 0.373 | 0.161 | 0.348 |
| | Setting 2 | 0.383 | 0.174 | 0.356 |
| | Setting 3 | 0.132 | 0.052 | 0.121 |

## A. ROUGE Scores

ROUGE (Recall Oriented Understudy for Gisting Evaluation) introduced by [23] is a metric that is used to measure the degree of similarity between the summary generated by the model and a reference summary, one that is usually written by humans. We utilised the `pyrouge` library in python to compute the degree of similarity between model generated summaries and our summaries. We report the ROUGE F1 scores in Table II. The best scores in each column are shown in bold, and the best score for each model corresponding to the three settings are underlined. From Table II, it is clear that the first model which utilized the HIPORank algorithm for extractive summarisation and LED model for subsequent abstractive summarization demonstrated the best performance across all the three metrics: ROUGE-1,2 and L scores. For 50% of the tests, the models performed best under setting-2, implying that the median value of 3 for no_repeat_ngram_size and num_beams= 2 for the beam search algorithm gave the best performance.

## B. N-Gram Abstractivess

We use the metric introduced by [24] to calculate the n-gram abstractiveness of the generated summaries by the model. This measures how many unique n-grams were created by the model during inference. The formula for n-gram abstractiveness as stated by the authors in the paper is calculated according to Equation (3)

$$1 - \frac{\text{\# of summary words part of n-gram copied}}{\text{total \# of summary words}} \quad (3)$$

The numerator is calculated by creating a list of n-grams in the source text and the generated summary. Next, we find the intersections between the two lists, then count the number of words in the intersection. To ensure that words that appear more than once in the source and generated summaries are not excluded from the intersection and are included with their correct count, we add an initial counting step to get the exact count of each word in the source and generated summaries.

This evaluation was carried out by generating summaries by the models for each of the settings and measuring the n-gram abstractiveness for n=1 to n=6, for all the generated summaries, then averaging the scores. We also calculated the n-gram abstractiveness for the ground truth summaries and plotted them for comparison. Fig. 5 (a-d) display the n-gram abstractiveness for each of the models. With this metric of evaluation, we obtained that the model HIPORank-LED consistently outperforms other models, going so far as to surpass the n-gram abstractiveness of the ground truth summaries for n $\geq$ 6.

## C. Human Evaluation

We asked 11 volunteers to rate the quality of our generated summaries. This step helps us to understand how well the

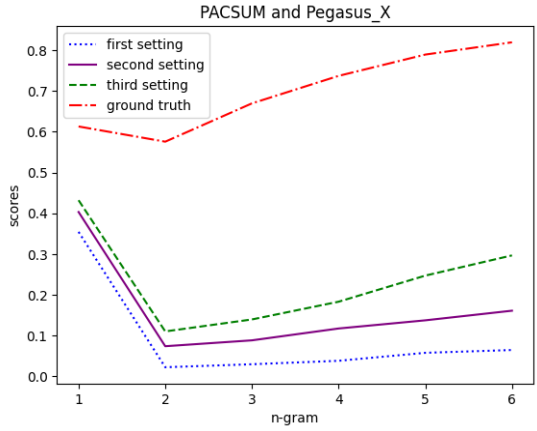

(a) n-gram abstractiveness for PacSum-PegasusX

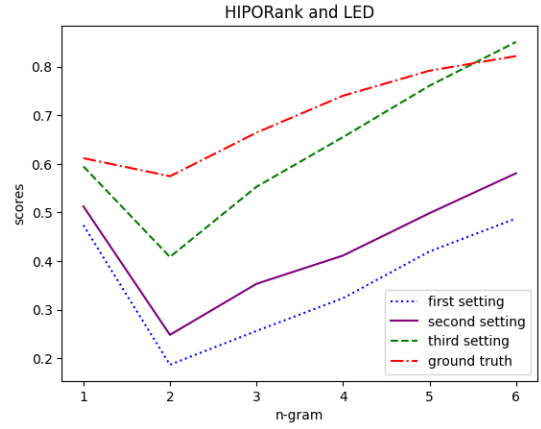

(b) n-gram abstractiveness for HIPORank-LED

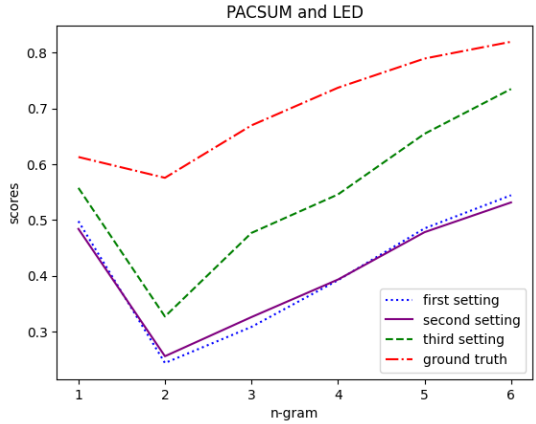

**(c)** n-gram abstractiveness for PacSum-LED

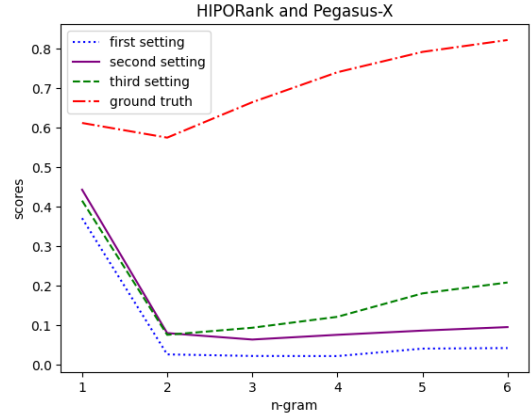

**(d)** n-gram abstractiveness for HIPORank-PegasusX

Fig. 5 N-gram abstractiveness for various models

summaries generated by the models are understood by human beings. Each volunteer has at least a bachelor's degree in Computer Engineering / Information Technology with sound understanding of the concepts of Natural Language Processing concepts. We sent the reviewers a source article and four summaries generated by the models, without sharing the names of the models that generated the summaries. We asked the reviewers to rate the summaries on a scale of 1-5, based on the following parameters:

**Fluency**: This relates to the grammatical quality of the generated summaries. It evaluates how well language flows in the text by measuring aspects such as the grammar, syntax and vocabulary.

**Coherence**: This metric evaluates how well the sentences or parts of text are connected together. It measures the logical organization of the sentences in the generated summary.

**Factuality**: This measures whether the facts stated in the summary text are consistent with the source text. This means that the facts stated in the summary should not be fabricated.

**Relevance:** This metric measures whether the generated summary is able to report the salient facts from the source text. It measures the relevance of the generated summary in capturing the key points of the source text

After getting their scores, we averaged the results, which are shown in Table III. From the results in Table III, we observe that the model PacSum-LED is the best performing model on all four parameters: Fluency, coherence, factuality and relevance.

TABLE III
RESULTS OF HUMAN EVALUATION

| Hyperparameter | Fluency | Coherence | Factuality | Relevance |
|---|---|---|---|---|
| PacSum-PegasusX | 4 | 3.72 | 4 | 4 |
| HIPORank-PegasusX | 4 | 3.63 | 3.72 | 3.54 |
| PacSum-LED | **4.27** | **4.45** | **4.27** | **4.27** |
| HIPORank-LED | 3.81 | 4.09 | 4.09 | 4.09 |

*D. Qualitative Evaluation*

To assess the quality of summaries, the reader is referred to Table IV, where we have provided the summary of a scientific paper generated by all the four models, and the ground truth summary for reference.

TABLE IV

REFERENCE SUMMARIES

| | Summary of: An Empirical Approach to Conceptual Case Frame Acquisition [25] |
|---|---|
| **Ground Truth** | **An empirical approach to conceptual case frame acquisition**<br>Conceptual natural language processing systems usually rely on case frame instantiation to recognize events and role objects in text. But generating a good set of case frames for a domain is time-consuming, tedious, and prone to errors of omission. We have developed a corpus-based algorithm for acquiring conceptual case frames empirically from unannotated text. Our algorithm builds on previous research on corpus-based methods for acquiring extraction patterns and semantic lexicons. Given extraction patterns and a semantic lexicon for a domain, our algorithm learns semantic preferences for each extraction pattern and merges the syntactically compatible patterns to produce multi-slot case frames with selectional restrictions. The case frames generate more cohesive output and produce fewer false hits than the original extraction patterns. Our system requires only pre-classified training texts and a few hours of manual review to filter the dictionaries, demonstrating that conceptual case frames can be acquired from unannotated text without special training resources. Our conceptual case frame acquisition project, extraction patterns, a domain semantic lexicon, and a list of conceptual roles and associated semantic categories for the domain are used to produce multiple-slot case frames with selectional restrictions. |
| **HIPORank-LED** | **A corpus-based approach to conceptual case frame acquisition**<br>Conceptual case frames are a valuable resource for information extraction systems that are designed for a domain specific purpose. We develop a new corpus based approach that learns semantic preferences for extraction patterns and merge these patterns into more complex structures. The conceptual cases frames can have slots for multiple role objects and each slot has a set of learned selectional restrictions for its role object. The case frames can be instantiated multiple times in a single structure to simplify subsequent discourse processing. In the past few years, several systems have been developed to generate structures of information extraction automatically. Unfortunately, acquiring a good set of case frames for a particular event can be a major undertaking. We present a corpus-based approach to the conceptual case framing acquisition that uses semantic preferences to merge syntactically compatible patterns into larger structures. We use an algorithm to generate conceptual case frame structures that are automatically extracted from a text corpus. we show several examples of cases frames that were generated automatically using this method. The semantic preferences are learned by learning semantic preferences and merging syntactical patterns into a larger structure. The resulting case frames contain slots for role objects that are associated with each event. |
| **PacSum-LED** | **A corpus-based approach to conceptual case frame acquisition**<br>Conceptual case frames are a fundamental component of natural language processing. They are typically lexically indexed so that each case frame can contain slots for thematic roles that are associated with each event. The conceptual cases frames should be more reliable at identifying relevant information (our experimental results support this hypothesis), and the case frames can instantiate multiple role objects in a single structure to simplify subsequent discourse processing. In the past few years, several systems have been developed to generate structures for information extraction automatically. Unfortunately, acquiring a good set of case frames for a domain can be a major undertaking. We present a new corpus based approach that uses the extraction patterns as a building block for constructing conceptual cases frame structures. Our approach uses semantic preferences for extraction patterns and merging syntactically compatible patterns into more complex structures. The resulting case frames have slots for multiple role object types. We use semantic preferences to merge syntactic patterns that are compatible with each other. We show that the use of semantic preferences in the extraction pattern generation process can improve the performance of our system. |
| **HIPORank-PegasusX** | Conceptual natural language processing typically involves case frame instantiation to recognize events and role objects in text. for example, an NLP system designed for a business domain might use case frames to recognize business activities such as mergers, acquisitions, or joint ventures. In the past few years, several systems have been developed to generate structures for information extraction automatically. We present a new corpus-based |
| **PacSum-PegasusX** | approach to conceptual case frame acquisition. We show that the conceptual frame frames can be automatically generated using unannotated text as input, coupled with a few hours of manual review. We then show several examples of case frames that were generated automatically using this method. The case frame would contain slots for thematic roles that are associated with each event. Our conceptual frame should be more reliable at identifying relevant information (our experimental results support this hypothesis), and the frame can instantiate multiple role object in a single structure to simplify subsequent discourse processing. Our new system constructs conceptual case frames by learning semantic preferences for extraction patterns and merging syntactically compatible patterns into more complex structures. The resulting case frames can have slots for multiple role objects and each slot has a set of learned selectional restrictions for its role object. Our empirical approach aims to exploit the text corpus to automatically acquire the syntactic and semantic role assignments that are necessary to achieve good performance in the domain. Finally, we present experimental results that compare the performance of the case frames with the extraction patterns.<br><br>Conceptual natural language processing typically involves case frame instantiation to recognize events and role objects in text. For example, an NLP system designed for a business domain might use case frames to recognize business activities such as mergers, acquisitions, or joint ventures. In the past few years, several systems have been developed to generate structures for information extraction automatically. our approach builds upon earlier work on corpus-based methods for generating extraction patterns and semantic lexicons. We have shown that conceptual case frames can be generated automatically using unannotated text as input, coupled with a few hours of manual review. Our empirical approach aims to exploit the text corpus to automatically acquire the syntactic and semantic role assignments that are necessary to achieve good performance in the domain. We propose a new corpus-based algorithm that uses the extracted patterns as a building block for constructing conceptual case frame structures. The resulting case frames are able to have slots for multiple role objects and each slot has a set of learned selectional restrictions for its role object. We show that the conceptual frame frames are more reliable at identifying relevant information (our experimental results support this hypothesis), and the case frame can instantiate multiple role object in a single structure to simplify subsequent discourse processing. Our new system constructs conceptual frame structures by learning semantic preferences for extraction patterns, and merging syntactically compatible patterns into more complex structures. We then present experimental results that compare the performance of the case frames with the extraction patterns. |

## VI. Conclusion and Future work

In this paper, we have performed experiments on summarization using a combination of two models for extractive and two models for abstractive summarization. We fine-tuned the pre-trained models on a dataset that is specific to summarization of long documents (in our case, scientific papers) and built a pipeline consisting of two major steps in summarization of those papers. We further performed inference on the created combination of models by using three evaluation metrics and have reported the results. We find that it is also possible to exceed the abstractiveness of ground-truth summaries using a certain combination of hyperparameters. Further work in this direction can be focused on applying techniques to improve the ROUGE-score and abstractiveness scores. Additionally, we would also like to perform experiments in extending this approach towards summarization of long documents in niche areas such as legal documents or biomedical data.